\documentclass[%
 aip,
 amsmath,amssymb,
 reprint,%
]{revtex4-1}

\usepackage{graphicx}
\usepackage{dcolumn}
\usepackage{bm}

\usepackage[utf8]{inputenc}
\usepackage[T1]{fontenc}
\usepackage{mathptmx}
\usepackage{etoolbox}
\usepackage{algorithm}
\usepackage{algorithmic}
\usepackage{amsmath}
\usepackage{multirow}
\usepackage{xcolor}
\usepackage{amsfonts}
\usepackage{subfigure}
\usepackage{float}
\usepackage[hidelinks]{hyperref} 
\hypersetup{
    colorlinks=true,
    linkcolor=blue,
    filecolor=blue,
    urlcolor=blue,
    citecolor=blue
}

\makeatletter
\def\@email#1#2{%
 \endgroup
 \patchcmd{\titleblock@produce}
  {\frontmatter@RRAPformat}
  {\frontmatter@RRAPformat{\produce@RRAP{*#1\href{mailto:#2}{#2}}}\frontmatter@RRAPformat}
  {}{}
}%
\makeatother
\begin{document}

\preprint{AIP/123-QED}

\title[SiMiC: Context-Aware Silicon Microstructure Characterization]{SiMiC: Context-Aware Silicon Microstructure Characterization Using Attention-Based Convolutional Neural Networks for Field-Emission Tip Analysis}

\author{Jing Jie Tan}
\email{tanjingjie@1utar.my}
\affiliation{ 
Department of Mechatronics and Biomedical Engineering, Lee Kong Chian Faculty of Engineering and Science, Universiti Tunku Abdul Rahman, Malaysia}%
\affiliation{%
Fakultät Angewandte Natur- und Kulturwissenschaften, Ostbayerische Technische Hochschule Regensburg, Germany
}%

\author{Rupert Schreiner}
\affiliation{%
Fakultät Angewandte Natur- und Kulturwissenschaften, Ostbayerische Technische Hochschule Regensburg, Germany
}%

\author{Matthias Hausladen}
\affiliation{%
Fakultät Angewandte Natur- und Kulturwissenschaften, Ostbayerische Technische Hochschule Regensburg, Germany
}%

\author{Ali Asgharzade}
\affiliation{%
Fakultät Angewandte Natur- und Kulturwissenschaften, Ostbayerische Technische Hochschule Regensburg, Germany
}%

\author{Simon Edler}
\affiliation{%
KETEK GmbH, Munich, Germany
}%

\author{Julian Bartsch}
\affiliation{%
KETEK GmbH, Munich, Germany
}%

\author{Michael Bachmann}
\affiliation{%
KETEK GmbH, Munich, Germany
}%

\author{Andreas Schels}
\affiliation{%
KETEK GmbH, Munich, Germany
}%

\author{Ban-Hoe Kwan}
\affiliation{ 
Department of Mechatronics and Biomedical Engineering, Lee Kong Chian Faculty of Engineering and Science, Universiti Tunku Abdul Rahman, Malaysia}

\author{Danny Wee-Kiat Ng}
\affiliation{ 
Department of Mechatronics and Biomedical Engineering, Lee Kong Chian Faculty of Engineering and Science, Universiti Tunku Abdul Rahman, Malaysia}

\author{Yan-Chai Hum}
\affiliation{ 
Department of Mechatronics and Biomedical Engineering, Lee Kong Chian Faculty of Engineering and Science, Universiti Tunku Abdul Rahman, Malaysia}

\date{\today}

\begin{abstract}
Accurate characterization of silicon microstructures is essential for advancing microscale fabrication, quality control, and device performance. Traditional analysis using Scanning Electron Microscopy (SEM) often requires labor-intensive, manual evaluation of feature geometry, limiting throughput and reproducibility. In this study, we propose SiMiC: Context-Aware Silicon Microstructure Characterization Using Attention-Based Convolutional Neural Networks for Field-Emission Tip Analysis. By leveraging deep learning, our approach efficiently extracts morphological features—such as size, shape, and apex curvature—from SEM images, significantly reducing human intervention while improving measurement consistency. A specialized dataset of silicon-based field-emitter tips was developed, and a customized CNN architecture incorporating attention mechanisms was trained for multi-class microstructure classification and dimensional prediction. Comparative analysis with classical image processing techniques demonstrates that SiMiC achieves high accuracy while maintaining interpretability. The proposed framework establishes a foundation for data-driven microstructure analysis directly linked to field-emission performance, opening avenues for correlating emitter geometry with emission behavior and guiding the design of optimized cold-cathode and SEM electron sources. The related dataset and algorithm repository that could serve as a baseline in this area can be found at \url{https://research.jingjietan.com/?q=SIMIC}.

\end{abstract}

\maketitle

\section{Introduction}

Accurate characterization of microstructures using scanning electron microscopy (SEM) is crucial for applications in fabrication and microscopy, as it enables precise analysis of material properties, surface morphology, field emission, and feature dimensions—essential for quality control, device performance, and the development of advanced microscale technologies \cite{Buchner2024,Bachmann2022,Singh2023}. Conventionally, tip geometry is obtained through direct SEM imaging or tip-characterizer methods, which are both time-consuming and reliant on manual analysis, limiting throughput and reproducibility \cite{Dai2020}. 

Recent advances in machine learning (ML) have introduced powerful tools to automate SEM image analysis and extract quantitative metrics of feature size, shape, and structure \cite{Bachmann2022}. ML and artificial intelligence (AI) techniques are increasingly integrated into electron microscopy workflows to overcome analytical bottlenecks and reduce human workload \cite{botifoll2022aiem}. In this context, deep learning models—particularly convolutional neural networks (CNNs)—have demonstrated remarkable performance in tasks such as object segmentation, particle-shape classification, and dimensional inference from microscopy images, enabling fast and objective microstructural characterization \cite{bals2023semseg}.

Building upon these developments, computer-vision models can now recognize and quantify fine silicon microstructural features with high precision~\cite{Kalkan2024}. Extending this methodology to field-emission applications, ML-based image analysis can be utilized to study emitter tips—such as tungsten field-emission gun (FEG) tips and carbon nanotube (CNT) cathodes—using SEM or transmission electron microscopy (TEM). These models can extract critical geometric parameters including apex radius, taper angle, and surface morphology, which directly influence the local field-enhancement factor as described by Fowler–Nordheim theory. The emission current is exponentially dependent on tip sharpness and surface uniformity~\cite{Borghei2024}. 

By providing precise, image-based measurements of these geometric features, ML-based predictions can serve as quantitative inputs for improved emission modeling or data-driven performance prediction. For example, CNN-derived tip radii can be incorporated into modified Fowler–Nordheim formulations or combined with machine-learning regressors ~\cite{Pradhan2025}. Integrating ML in this manner enables automated inspection, classification, and optimization of emitter geometries, reducing reliance on manual SEM analysis~\cite{Li2025}. 

Overall, this synergy between deep-learning-based microstructure analysis and field-emission modeling establishes a microstructure-informed pathway toward predictive emitter engineering. It promises accelerated design cycles for high-performance cold cathodes and field-emission electron sources, bridging the gap between nanoscale morphology and macroscopic emission behavior~\cite{Borghei2024}.

\begin{table*}
\caption{Typical Resolution and Influencing Factors in Various Microscopy Techniques}
\centering
\begin{tabular}{|p{2.5cm}| p{2.5cm}| p{12cm}|}
\hline
\textbf{Microscopy} & \textbf{Error Interval} & \textbf{Influencing Factors} \\
\hline
\textbf{Scanning Electron Microscopy (SEM)} &  $\sim ~2-10 nm$ & Electron beam spot size and stability, magnification calibration, detector noise, sample charging or drift, vacuum quality, and environmental stability. Proper calibration using traceable standards and maintaining a stable working distance are critical. \cite{chiang2024discussion}\\
\hline
\textbf{Transmission Electron Microscopy (TEM)} & $\sim0.05-0.5nm$ & Limited by objective lens aberrations (mitigated by Cs-correction), sample thickness and preparation quality, beam-induced damage or shrinkage, and mechanical vibration or drift. Requires electron-transparent specimens. \cite{thermofisher_sem_tem_difference}\\
\hline
\textbf{Atomic Force Microscopy (AFM)} & $\sim 0.1--0.5nm$ & Influenced by cantilever calibration (spring constant), tip radius and wear, scanner (piezo) calibration, feedback control, sample-tip interactions, thermal drift, and environmental conditions (vibration, humidity). Lateral accuracy is also affected by tip convolution and surface compliance. \cite{brukerAtomicForce}\\
\hline
\textbf{Optical Microscopy} & $\sim 200-300nm$& Governed by the numerical aperture and illumination wavelength (diffraction limit), refractive index of the medium, optical aberrations, pixel resolution, scanning accuracy, and stage stability. Super-resolution techniques can achieve $\sim$10--50\,nm lateral resolution under optimal conditions. \cite{microscopyuDiffractionBarrier}\\
\hline
\end{tabular}
\label{tab:microscopy-resolution}
\end{table*}

\section{Literature Review}

\subsection{Silicon Microstructure Measurement}
In the characterization of silicon microstructures, high-throughput sample analysis is often constrained by the time-intensive nature of conventional measurement techniques. When dealing with large volumes of data or numerous samples, relying solely on physical measurements can become inefficient and resource-heavy. ML methods present a promising alternative for accelerating analysis and could identify underlying patterns for other field applications \cite{Tan2025psychonlu}. For ML-based predictions to be acceptable in place of physical measurements, their uncertainty must match or fall below the inherent resolution and error margins of the instrumentation being replaced. Table~\ref{tab:microscopy-resolution} summarizes the typical resolution and primary influencing factors associated with several widely used microscopy techniques in microstructure characterization. This provides a benchmark for evaluating the accuracy requirements that ML models must meet in order to serve as viable surrogates.

\subsection{Classical Computational Approaches}
Classical computational image processing techniques are fundamental for analyzing silicon microstructures from SEM images, which is crucial for understanding material properties in applications like photovoltaics and microelectronics \cite{Horstmann2012}. SEM provides high-resolution images of silicon's surface topography and composition, enabling detailed microstructural characterization \cite{Kuchar2023}. Core techniques include pre-processing for noise reduction and contrast enhancement, thresholding to segment distinct phases, pores, and grain boundaries (often after chemical etching), edge detection (e.g., Sobel, Canny) to delineate feature boundaries, and morphological operations (dilation, erosion, opening, closing) to refine segmentation, separate touching features, and connect broken boundaries \cite{Ali2023}. Contour tracing algorithms precisely delineate grain and pore boundaries, enabling the quantification of grain size distributions, shape factors (like aspect ratio or circularity), and orientation relationships \cite{Vippola2016}. These deterministic approaches offer significant advantages due to their interpretability, reproducibility, and relatively low computational cost, facilitating semi-automated workflows that reduce manual labor and subjectivity in silicon microstructure characterization.

\subsection{Machine Learning Approaches}
Machine learning has been increasingly applied to SEM and related electron microscopy images for object detection, segmentation, and quantitative analysis \cite{Continuum2024}. In general, ML can detect and identify objects of high variability in SEM images more rapidly than manual methods \cite{Bangaru2022}. For example, Bals et al.\cite{Bals2023} used deep CNNs (U-Net architectures) to segment nanoparticles in SEM micrographs, separating particles from background and classifying their shapes. Other studies have employed object-detection networks on SEM data. López-Gutiérrez et al.\cite{lopez2022yolo} trained YOLO models on SEM images of silver nanoparticles, achieving robust detection even with synthetic training images. Although proximity of particles posed challenges, YOLO architectures proved fast and accurate. These works illustrate that both segmentation (e.g., U-Net) and detection frameworks (e.g., YOLO) have been successfully applied to SEM imagery. General reviews further confirm that ML in electron microscopy can automate tasks from feature recognition to spectroscopic mapping \cite{botifoll2022aiem}. To the best of our knowledge, we haven't found any previous work specifically addressing silicon microstructures; hence, we established a dataset with a proposed classification algorithm. 

\section{Methodology}

\begin{figure*}
    \centering
    \includegraphics[width=1\linewidth]{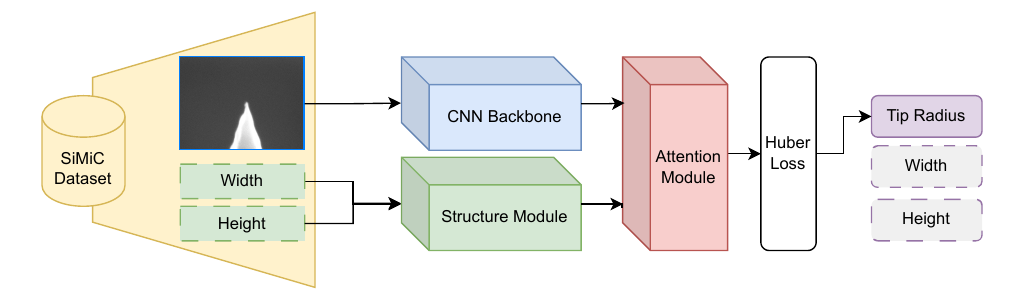} 
    \caption{Overview of the model architecture. The CNN backbone (blue rectangle) extracts features from the input image to predict target parameters. The optional structure module (green rectangle) provides reference dimensions to align the output, while the attention module (red rectangle) facilitates interaction between the backbone and structure features.}
    \label{fig:arch}
\end{figure*}

The overall model architecture and training procedure are illustrated in Figure \ref{fig:arch}, beginning with dataset processing, followed by the machine learning algorithm, and concluding with the loss computation and output for evaluation.

\subsection{Dataset Preparation}

A dataset consisting of 900 samples was obtained using a Scanning Electron Microscope (SEM). The SEM was operated under carefully controlled imaging conditions to ensure consistent sample quality across the entire dataset. These imaging parameters were selected to achieve sufficient surface detail for downstream computational analysis, particularly for tasks such as automated feature extraction, computer vision processing, and model-driven characterization. The parameters used during data acquisition are summarized in Table~\ref{tab:sem_params}. 

\begin{table}[H]
\centering
\caption{SEM Imaging Parameters Used to Generate the Dataset}
\label{tab:sem_params}
\begin{tabular}{ll}
\hline
\textbf{Parameter} & \textbf{Value} \\
\hline
Beam current & 300 pA \\
Acceleration voltage & 5 kV \\
Working distance & 10 mm \\
Sample tilt & 45$^\circ$ \\
Field of view (tip radii measurement) & 1 µm \\
Resolution & 1024 $\times$ 768 \\
\hline
\end{tabular}
\end{table}

Following acquisition, the dataset was prepared for machine learning model development. To ensure a robust assessment of model performance, the dataset was partitioned into three distinct subsets: training, validation, and evaluation. Initially, the dataset was split into training and evaluation sets in an 80:20 ratio, thereby reserving 20\% of the data as a completely unseen evaluation set to be used solely for final performance assessment. Subsequently, the training subset, which accounts for 80\% of the total data, was further divided into training and validation sets, again following an 80:20 split. The resulting validation set was utilized exclusively for hyperparameter tuning, facilitating model optimization without compromising the integrity of the final evaluation process.

\begin{figure*}
    \centering
    \includegraphics[width=1\linewidth]{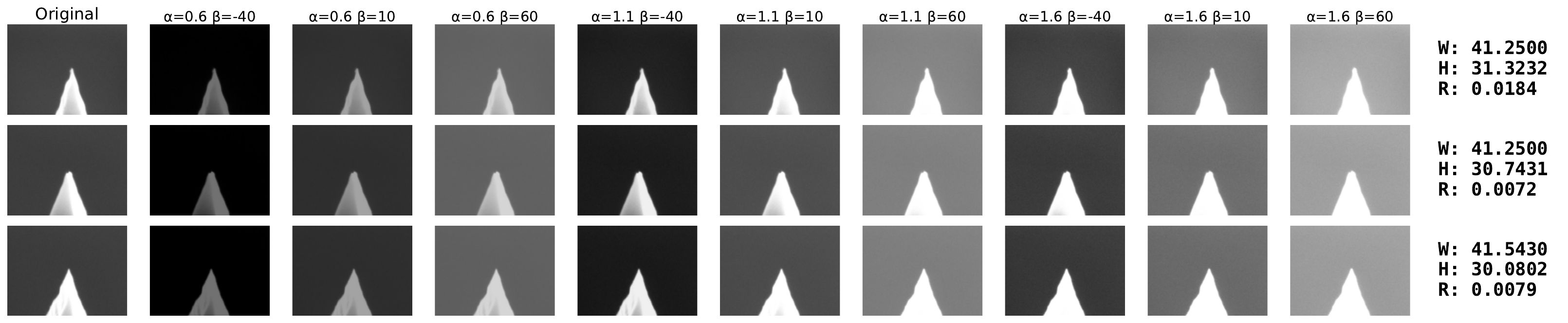}
    \caption{The figure presents the original images alongside their augmented versions, generated through systematic variations in brightness ($\beta = -40, 10, 60$) and contrast ($\alpha = 0.6, 1.1, 1.6$), including the width (W), height (H), and radius (R). These augmentations aim to simulate varying lighting conditions due to different exposure levels and adjustments by SEM.}
    \label{fig:augment}
\end{figure*}

To further improve model performance, the dataset was augmented by varying the contrast and brightness of the images. Since scanning electron microscope (SEM) images often exhibit substantial variability in these parameters, each original image $I$ was used to generate nine augmented variants denoted as $\{ I_k \}_{k=1}^9$. As defined upon first appearance, $I_k$ represents the $k^\text{th}$ augmented image, $\alpha_k$ is the contrast‐adjustment factor, and $\beta_k$ is the brightness offset, which together form the linear transformation in Eq.~\ref{eq:contrast_brighness}. Specifically, the augmented image is computed as $I_k = \alpha_k I + \beta_k$, ensuring that each adjustment is mathematically explicit. During training, every augmented image $I_k$ inherits the same label as its corresponding original image $I$, maintaining label consistency while expanding the effective size of the dataset. Examples of the applied augmentations are shown in Fig.~\ref{fig:augment}.

\begin{equation}
I_k = \alpha_k \cdot I + \beta_k, \quad k = 1, 2, \ldots, 9,
\label{eq:contrast_brighness}
\end{equation}

\subsection{SiMiC Machine Learning Model}

As shown in Figure~\ref{fig:arch}, it consists of a Convolutional Neural Network (CNN) backbone that processes the input image to predict parameters such as width, height, and radius. An optional structure module can be included to provide reference information about the input dimensions, which helps to align the output shape. This alignment enables the attention mechanism to focus more effectively on relevant features during training. Additionally, an attention module is employed to coordinate and integrate features from both the backbone and the structure module.

\subsubsection{Convolutional Neural Network Backbone}
In our exploration of deep learning model performance, we evaluated three prominent convolutional neural network (CNN) architectures: ResNet18, EfficientNet, and MobileNet. Each of these architectures offers distinct advantages in terms of computational efficiency, scalability, and accuracy, making them suitable for different application scenarios.

\textbf{ResNet18} \cite{he2016deep,he2016identity} is a widely adopted CNN model that incorporates residual learning through the use of skip connections. These skip connections help mitigate the vanishing gradient problem that commonly arises in deep neural networks, thereby facilitating the training of deeper architectures. ResNet18 strikes a balance between model complexity and performance, making it a reliable choice for a wide range of computer vision tasks.

\textbf{EfficientNet} \cite{tan2019efficientnet} introduces a novel scaling method known as compound scaling, which simultaneously scales depth, width, and input resolution in a principled manner. This approach enables EfficientNet to achieve state-of-the-art performance while maintaining a significantly lower parameter count compared to conventional models. Its design ensures high accuracy with optimal computational efficiency, making it suitable for both research and production environments.

\textbf{MobileNet} \cite{mobilenet} is specifically designed for deployment on mobile and embedded devices with limited computational resources. It employs depthwise separable convolutions, which decompose standard convolutions into a depthwise and a pointwise convolution. This significantly reduces the model's computational cost and parameter count without a substantial loss in accuracy. MobileNet is thus an excellent choice for real-time applications on resource-constrained platforms.

\subsubsection{Structure Module}

To enhance geometric reasoning in convolutional neural networks (CNNs) while preserving compatibility with transfer learning, we adopt a CoordConv-based design. This method enriches the input tensor by embedding explicit spatial coordinates, allowing the network to better infer geometric properties (such as tip radius) from visual patterns alone.

In our model architecture, we introduce a mechanism to incorporate object-specific geometric information—specifically width and height—into the attention mechanism. This is achieved through a learnable linear projection layer, defined as: $self.embed_{wh} = \texttt{nn.Linear}(2, d)$. This layer takes a 2-dimensional vector representing the width and height of the object and projects it into a $d$-dimensional embedding space. This geometric embedding serves as a high-level descriptor that guides the attention mechanism to focus on spatial regions in the image feature map that are relevant to the given object dimensions. This design enables the model to attend adaptively based on the scale and shape of the target object, improving the precision of downstream regression tasks, such as radius prediction.

\subsubsection{Attention Modules}

This module is optional and is applied when incorporating structural information such as object width and height. To enhance feature integration between the CNN backbone and the structure module (see Figure~\ref{fig:arch}), we investigate two distinct attention mechanisms. Specifically, to model the interaction between spatial features and object scale, we experiment with two attention mechanisms: Bahdanau-style additive attention and Multi-Head Attention (MHA).

Given the structure vector $\mathbf{S}_i \in \mathbb{R}^2$, we first project it into an embedding space to obtain the query vector:

\begin{equation}
    \mathbf{q}_i = \text{Linear}_{\mathrm{struct}}(\mathbf{S}_i) \in \mathbb{R}^d,
    \label{eq:query_vector}
\end{equation}
where each symbol is defined as follows:
\begin{itemize}
    \item $\mathbf{S}_i \in \mathbb{R}^2$: Structure vector representing width and height.
    \item $\text{Linear}_{\mathrm{struct}}$: Learnable linear projection layer for the structure vector.
    \item $\mathbf{q}_i \in \mathbb{R}^d$: Query vector derived from the structure vector.
    \item $d$: Embedding dimension.
\end{itemize}

The CNN feature map is flattened to obtain $N$ spatial positions, each projected into key-value pairs. For additive attention, these are denoted as $\{\mathbf{k}_{ij}, \mathbf{v}_{ij}\}$, while for MHA, the keys and values are arranged into matrices $\mathbf{K}_i, \mathbf{V}_i \in \mathbb{R}^{N \times d}$.

\paragraph{Additive Attention \cite{bahdanau2015neural}}
As in Eq. \ref{eq:additive_attention} and Eq. \ref{eq:additive_output} computes attention scores by applying a feed-forward network with a learnable context vector to combined query and key representations, enabling the model to focus on relevant parts of the input sequence dynamically. 

\begin{equation}
    \alpha_{ij} = \frac{\exp\left(\mathbf{u}^\top \tanh(\mathbf{W}_q \mathbf{q}_i + \mathbf{W}_k \mathbf{k}_{ij})\right)}{\sum_{j'=1}^{N} \exp\left(\mathbf{u}^\top \tanh(\mathbf{W}_q \mathbf{q}_i + \mathbf{W}_k \mathbf{k}_{ij'})\right)}
    \label{eq:additive_attention}
\end{equation}

\begin{equation}
    \mathbf{z}_i = \sum_{j=1}^{N} \alpha_{ij} \mathbf{v}_{ij}
    \label{eq:additive_output}
\end{equation}
where each symbol is defined as follows:
\begin{itemize}
    \item $\mathbf{k}_{ij}, \mathbf{v}_{ij} \in \mathbb{R}^d$: Key and value vectors at position $j$ for sample $i$.
    \item $\mathbf{W}_q, \mathbf{W}_k \in \mathbb{R}^{d \times d}$: Learnable weight matrices.
    \item $\mathbf{u} \in \mathbb{R}^d$: Learnable context vector.
    \item $\alpha_{ij}$: Attention weight for position $j$.
    \item $\mathbf{z}_i \in \mathbb{R}^d$: Final attended feature vector.
\end{itemize}

\paragraph{Multi-Head Attention \cite{vaswani2017}}
The attention mechanism captures complex feature-scale interactions using scaled dot-product attention. For head $i$, the output vector $\mathbf{z}_i$ is computed as in Equation \ref{eq:mha_output}

\begin{equation}
    \mathbf{z}_i = \text{softmax}\Bigg( \frac{\mathbf{q}_i \mathbf{K}_i^\top}{\sqrt{d}} \Bigg) \mathbf{V}_i \in \mathbb{R}^d,
    \label{eq:mha_output}
\end{equation}

where each symbol is defined as follows:

\begin{itemize}
    \item $\mathbf{q}_i \in \mathbb{R}^{1 \times d}$: Query vector for head $i$.
    \item $\mathbf{K}_i \in \mathbb{R}^{N \times d}$: Matrix of key vectors for $N$ spatial positions.
    \item $\mathbf{V}_i \in \mathbb{R}^{N \times d}$: Matrix of value vectors corresponding to the keys.
    \item $d$: Dimensionality of the query, key, and value vectors.
    \item $\mathbf{z}_i \in \mathbb{R}^d$: Output vector of the scaled attention for head $i$.
\end{itemize}

\subsection{Training Loss}

For training and validation, we adopted the \textbf{Huber Loss}\cite{gokcesu, Meyer_2021_CVPR} (as in Eq. \ref{eq:huber}, which combines the robustness of the L1 loss with the smoothness of the L2 loss. This makes it less sensitive to outliers while remaining differentiable and stable for gradient-based optimization.

\begin{equation}
\label{eq:huber}
\mathcal{L}_{\text{Huber}} = \sum_{i=1}^{N} \ell_\delta(y_i, \hat{y}_i) = \sum_{i=1}^{N} 
\begin{cases}
\frac{1}{2\delta}(y_i - \hat{y}_i)^2, & \text{if } |y_i - \hat{y}_i| \leq \delta, \\
|y_i - \hat{y}_i| - \frac{\delta}{2}, & \text{otherwise,}
\end{cases}
\end{equation}

where each symbol is defined as follows:

\begin{itemize}
    \item $N$: Number of samples.
    \item $y_i$: Ground-truth value for the $i$-th sample.
    \item $\hat{y}_i$: Predicted value for the $i$-th sample.
    \item $\delta$: Threshold parameter that controls the transition between L1 and L2 behavior.
    \item $\ell_\delta(y_i, \hat{y}_i)$: Huber loss for the $i$-th sample.
\end{itemize}

\subsection{Experiment Setting}
The detailed hyperparameters used in our training procedure are summarized in Table \ref{tab:hyperparameters}. We adopted several deep learning models and applied data augmentation as well as attention mechanisms to enable comparison and establish a benchmark, serving as the first study on silicon microstructure characterization for field-emission tip analysis.

\begin{table}[H]
\centering
\caption{Training hyperparameters used in the experiments.}
\begin{tabular}{ll}
\hline
\textbf{Hyperparameter} & \textbf{Value} \\
\hline
Learning rate & $1\times10^{-4}$ \\
Batch size & 32 \\
Number of epochs & 500 (with early stopping) \\
Optimizer & ADAM \\
Weight decay & $1\times10^{-5}$ \\
Training hardware & NVIDIA A100 GPU \\
\hline
\end{tabular}
\label{tab:hyperparameters}
\end{table}

\subsection{Evaluation Metrics}

For evaluation, we use the \textbf{Root Mean Square Error (RMSE)} (Eq. \ref{eq:rmse}), which measures the average magnitude of prediction errors

\begin{equation}
\label{eq:rmse}
\text{RMSE} = \sqrt{\frac{1}{N} \sum_{i=1}^{N} (y_i - \hat{y}_i)^2},
\end{equation}

where each symbol is defined as follows:

\begin{itemize}
    \item $N$: Number of samples.
    \item $y_i$: Ground-truth value for the $i$-th sample.
    \item $\hat{y}_i$: Predicted value for the $i$-th sample.
\end{itemize}

Moreover, we evaluate the model using the \textbf{coefficient of determination (R\textsuperscript{2})} (Eq. \ref{eq:r2}), which measures the strength of correlation.

\begin{equation}
R^2 = 1 - \frac{\sum_{i=1}^{N} (y_i - \hat{y}_i)^2}{\sum_{i=1}^{N} (y_i - \bar{y})^2},
\label{eq:r2}
\end{equation}

where each symbol is defined as follows:

\begin{itemize}
    \item $N$: Number of samples.
    \item $y_i$: Ground-truth value for the $i$-th sample.
    \item $\hat{y}_i$: Predicted value for the $i$-th sample.
    \item $\bar{y} = \frac{1}{N}\sum_{i=1}^{N} y_i$: Mean of the ground-truth values.
\end{itemize}

Hence, RMSE provides a measure of the model's prediction error magnitude, while R\textsuperscript{2} evaluates how well the predictions explain the variance in the data, enabling future researchers to comprehensively assess model performance.

\section{Result and Discussion}

The experimental results, as shown in Table~\ref{tab:results}, reveal several key observations regarding the performance of different model configurations.

\begin{table*}
\center
\caption{Prediction performance for width, height, and radius was evaluated using different deep learning model architectures and attention mechanisms. Results are reported for both the full-prediction and half-prediction settings. In the half-prediction setting, width and height were provided as additional inputs for radius prediction, so only radius results are included. The table highlights the improvements achieved through multi-head attention and data-augmentation techniques.}
\begin{tabular}{|l|ccc|c|ccc|c|}
\hline
\multicolumn{1}{|c|}{\multirow{3}[0]{*}{\textbf{Approaches}}} & \multicolumn{4}{|c|}{\textbf{RMSE}} & \multicolumn{4}{|c|}{\textbf{R\textsuperscript{2}}} \\
\cline{2-9}
 &\multicolumn{3}{|c|}{\textbf{Full Prediction}} & \multicolumn{1}{|c|}{\textbf{Half Prediction}}  &\multicolumn{3}{|c|}{\textbf{Full Prediction}} & \multicolumn{1}{|c|}{\textbf{Half Prediction}} \\
\cline{2-9}
  & \multicolumn{1}{c}{\textbf{Width}} & \multicolumn{1}{c}{\textbf{Height}} & \multicolumn{1}{c|}{\textbf{Radius}} &\multicolumn{1}{c|}{\textbf{Radius}} & \multicolumn{1}{c}{\textbf{Width}} & \multicolumn{1}{c}{\textbf{Height}} & \multicolumn{1}{c|}{\textbf{Radius}} & \multicolumn{1}{c|}{\textbf{Radius}} \\
  \hline
\textbf{Resnet} & 1.1947 & 1.2817 & 0.0443 & 0.0225 & 0.2147& 0.2131& 0.2227 & 0.2623\\
EfficientNet & 1.2982 & 1.3182 & 0.0942 & 0.0313 & 0.2033 & 0.2189& 0.2293&0.2895 \\
MobileNet & 1.2712 & 1.3276 & 0.0574 & 0.0274 & 0.2113& 0.2221& 0.2231 & 0.2934\\
\hline
Additive Attention + Resnet & 1.1892 & 1.2801 & 0.0421 & 0.0221 & 0.2138&0.2112 &0.2234 &0.2908 \\
\textbf{Multihead Attention + Resnet} & 1.0097 & 1.2484 & 0.0395 & 0.0192 & 0.2259&0.2228 &0.2268 &0.2887 \\
\textbf{Augmentation + Multihead Attention + Resnet} & 0.9312 & 1.1158 & 0.0319 & 0.0117 & 0.2295 & 0.2330 &0.2362 & 0.3098 \\
\hline
\end{tabular}%

\label{tab:results}
\end{table*}

\subsection{Model Performance}
\textbf{Backbone Comparison:} Across the evaluated backbone architectures, ResNet consistently outperformed both EfficientNet and MobileNet in terms of RMSE for all predicted variables. For example, ResNet achieved lower errors in width (1.1947 vs. 1.2982 and 1.2712) and radius (0.0443 vs. 0.0942 and 0.0574), indicating a stronger ability to capture spatial features relevant to the task. Despite these improvements in RMSE, the R\textsuperscript{2}) values remained relatively limited possitive correlation (approximately 0.2) across all backbones. This limited coefficient of determination suggests that, although ResNet reduces prediction error, the model still struggles to explain a substantial proportion of the variance in the target variables. In other words, the current modeling setup exhibits a limited capacity to learn or reveal meaningful correlations between the input images and the predicted morphological parameters.

\textbf{Attention Mechanisms:} The integration of attention mechanisms improved the model's predictive performance. Among the evaluated approaches, multi-head attention outperformed both additive attention and the non-attention baseline. When paired with ResNet, multi-head attention reduced the radius prediction RMSE from 0.0443 (vanila ResNet) and 0.0421 (Additive Attention + ResNet) to 0.0395. This error was further reduced to 0.0319 when multi-head attention was combined with data augmentation, demonstrating a complementary effect between architectural enhancements and training diversity. A similar trend was observed in the coefficient of determination (R\textsuperscript{2}). The baseline ResNet achieved an R\textsuperscript{2} of 0.2323, which slightly decreased to 0.2234 with additive attention but improved to 0.2268 with multi-head attention. With the inclusion of data augmentation, the R\textsuperscript{2} further increased to 0.2362, the highest across all configurations, indicating that the combined strategy not only reduced absolute prediction error but also improved the model’s ability to explain variance in the target variable.

\textbf{Data Augmentation:} Applying data augmentation led to substantial gains in prediction accuracy. For instance, the full-prediction RMSE for radius decreased from 0.0395 (Multi-head Attention + ResNet) to 0.0319 when augmentation was incorporated. A similar improvement was observed in the half-prediction setting, where RMSE dropped from 0.0192 to 0.0117, highlighting the advantage of increased data diversity in enhancing model generalization. This trend was also reflected in the R\textsuperscript{2} values. In the full-prediction configuration, R\textsuperscript{2} improved from 0.2268 to 0.2362 with augmentation, while in the half-prediction scenario it increased from 0.2887 to 0.3098. These gains demonstrate that data augmentation not only lowers prediction error but also enhances the model’s ability to capture variance in the target variable.

\textbf{Inclusion of Width and Height as Features:} 
Incorporating width and height as additional input features in the half-prediction setting produced a substantial improvement in radius prediction accuracy. As shown in Table~\ref{tab:results}, the radius RMSE under half prediction is nearly halved across all model variants. For example, the ResNet radius RMSE decreased from 0.0443 to 0.0225, accompanied by an R\textsuperscript{2} increase from 0.2227 to 0.2623. Similarly, the augmented multi-head attention configuration saw its RMSE reduced from 0.0319 to 0.0117, with a corresponding R\textsuperscript{2} improvement from 0.2362 to 0.3098. This performance gain—approaching a 50\% reduction in error in several cases—highlights the strong predictive relationship between radius and the associated width and height, motivating further investigation into cross-feature interactions and their role in enhancing model performance.

\subsection{Feature Analysis}
\begin{figure*}

\subfigure[Half Prediction (Given Height and Width)]{
        \includegraphics[width=0.7\textwidth]{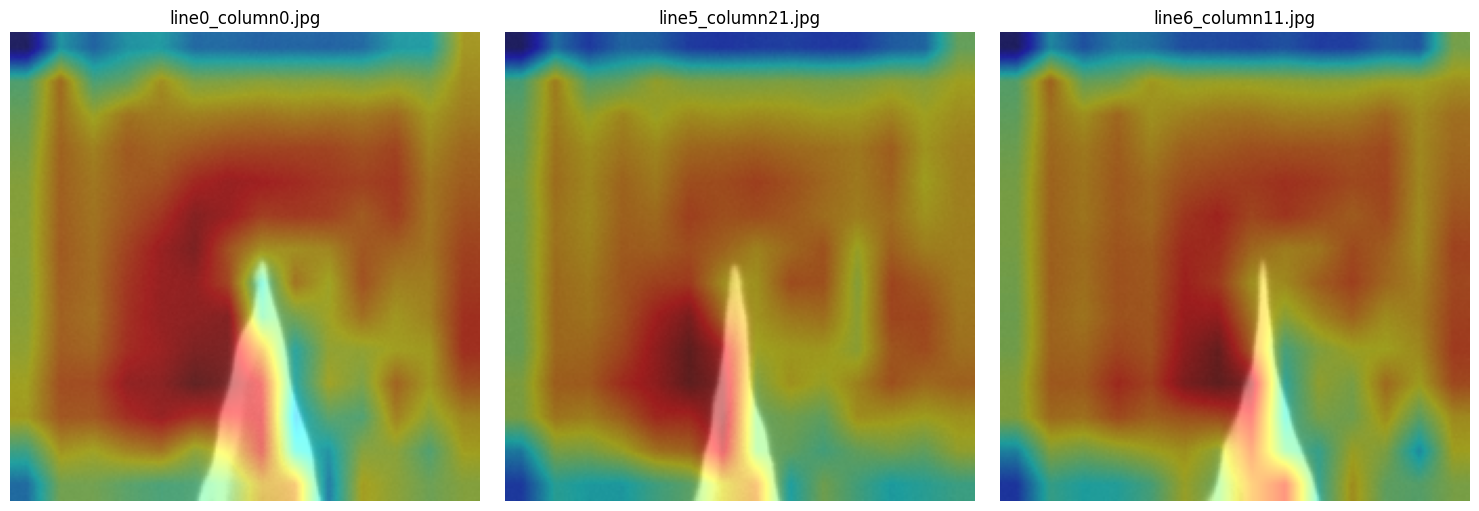}
        \label{fig:nohw}
    }
    \subfigure[Full Prediction (No Height and Width Given)]{
        \includegraphics[width=0.7\textwidth]{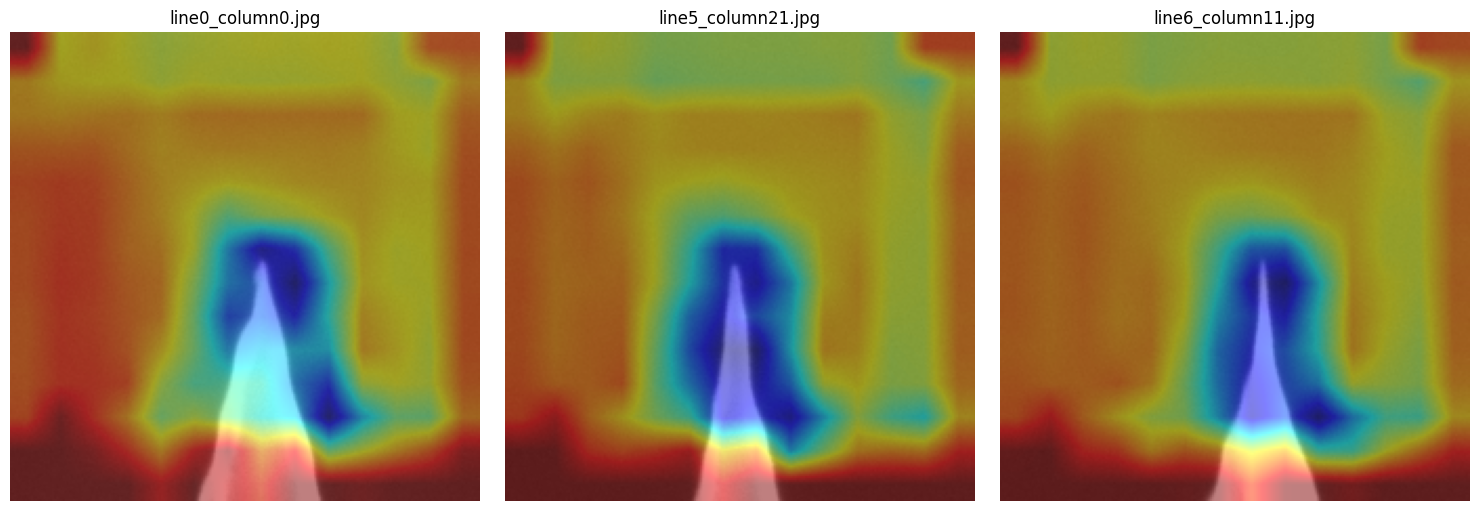}
        \label{fig:hw}
    }
    \caption{Attention map visualizations comparing two ResNet-based prediction settings: (a) Half Prediction, where object height and width are provided as input features, and (b) Full Prediction, where these geometric cues are absent. Red regions indicate high attention weights, while blue areas reflect minimal focus. When height and width are omitted, the model allocates greater attention to the object's bounding structure, implicitly inferring geometric attributes. Conversely, providing these features leads to more localized attention on object contours, improving radius estimation. Notably, background areas also receive attention, suggesting their utility as contextual anchors for object scale and positioning. The sample images were randomly picked from the dataset for visualization.}
    
    \label{fig:feature}
\end{figure*}

As in Figure \ref{fig:feature}, an analysis of the feature importance maps reveals insightful patterns in the model's attention behavior. 

In the visualization, regions highlighted in red correspond to high attention weights, while blue areas indicate minimal focus. When width and height are not provided as input features—and no additional spatial cues such as rotation or distance are available—the ResNet model appears to allocate substantial attention to estimating these missing geometric attributes. This is evident from the increased focus on the object's bounding structure, as the network implicitly attempts to infer width and height information from visual cues.

Conversely, when width and height are explicitly included as input features, the model gains a preliminary geometric understanding of the sample's orientation within 3D space. As a result, the attention shifts from estimating global dimensions to more localized analysis of object contours. This shift enables the network to refine its focus toward regions most informative for radius prediction, thus enhancing its accuracy.

Interestingly, in several cases, the model also appears to focus on background regions. Rather than being noise, these regions may serve as contextual references that aid the model in understanding relative object positioning or scale. This behavior suggests that the model leverages not only object-specific features but also environmental cues to inform its predictions.

\section{Conclusion}

In this study, we proposed a SiMiC approach and systematically evaluated the performance of various image model backbones as well as the attention mechanisms for the prediction of object dimensions, specifically width, height, and radius. Our findings demonstrate that ResNet serves as a robust backbone, outperforming other architectures such as EfficientNet and MobileNet. The incorporation of multi-head attention significantly enhances model performance beyond additive attention and non-attentive baselines. Furthermore, data augmentation consistently improves prediction accuracy, underscoring the importance of diverse training samples.

A key insight from our experiments is the substantial benefit of including width and height as explicit input features when predicting radius. This incorporation leads to nearly a 50\% reduction in radius prediction error, emphasizing the interdependence of geometric attributes and the model’s ability to leverage contextual information. Additionally, analysis of feature importance maps suggests that providing these features enables the model to better focus on relevant object contours rather than inferring missing dimensions, which in turn improves prediction precision.

To the best of our knowledge, there is no previous work specifically on silicon microstructure characterization, particularly regarding the tip geometry, using machine learning, and no publicly available dataset exists for this purpose. Therefore, we created a dataset and algorithm repository that could serve as a baseline in this area: \url{https://research.jingjietan.com/?q=SIMIC}.

However, it is important to note that, in the absence of prior work for direct comparison, RMSE primarily reflects the magnitude of prediction errors, while R\textsuperscript{2} provides insight into the strength of the correlation between predictions and true values. We found that even the best-performing model exhibits only a modest positive correlation. In future work, this framework could enable field-emission studies by correlating microstructural parameters with emission performance, allowing data-driven optimization of emitter tip geometry and material design. Technically, we could explore architectures inspired by visual transformer and decoder systems, such as those in \cite{Tan2024sia,Tan2025picepr}, potentially adapting them for classification tasks. Furthermore, approaches addressing hard samples and class imbalance, as discussed in \cite{Tan2025aflps}, could be investigated to enhance model robustness.

\begin{acknowledgments}
The authors sincerely thank KETEK for their valuable support and coordination in providing the sample. The authors gratefully acknowledge the support of the Deutscher Akademischer Austauschdienst (DAAD, German Academic Exchange Service) Research Grant in Germany, which facilitated the research collaboration between Universiti Tunku Abdul Rahman and Ostbayerische Technische Hochschule (OTH) Regensburg, enabling the mobility of the first author, Jing Jie Tan.
\end{acknowledgments}

\section*{Conflict of Interest}
The authors have no conflicts to disclose.

\section*{Data Availability Statement}
The data that support the findings of this study are openly available in the site: \url{https://research.
jingjietan.com/?q=SIMIC}.

\nocite{*}
\bibliography{references}

\end{document}